\documentclass[conference]{acmart}

\AtBeginDocument{%
  }

\usepackage{makecell}
\usepackage{multirow}

\usepackage{caption}
\usepackage{subcaption}

\begin{document}

\title{Exploring Interactions and Regulations in Collaborative Learning: An Interdisciplinary Multimodal Dataset}

\author{Yante Li}
\email{yante.li@oulu.fi}

\author{Yang Liu}
\email{yang.liu@oulu.fi}
\author{Khánh Nguyen}
\email{Andy.Nguyen@oulu.fi}
\author{Henglin Shi}
\email{henglin.shi@oulu.fi}
\author{Eija Vuorenmaa}
\email{eija.vuorenmaa@oulu.fi}
\author{Sanna Jarvela}
\email{sanna.jarvela@oulu.fi}
\author{Guoying Zhao}
\email{guoying.zhao@oulu.fi}
\affiliation{%
  \institution{University of Oulu}
  \city{Oulu}
  \country{Finland}
}

\renewcommand{\shortauthors}{}

\begin{abstract}
 Collaborative learning is an educational approach that enhances learning through shared goals and working together. Interaction and regulation are two essential factors related to the success of collaborative learning. Since the information from various modalities can reflect the quality of collaboration, a new multimodal dataset with cognitive and emotional triggers is introduced in this paper to explore how regulations affect interactions during the collaborative process. Specifically, a learning task with intentional interventions is designed and assigned to high school students aged 15 years old (N=81) in average. Multimodal signals, including video, Kinect, audio, and physiological data, are collected and exploited to study regulations in collaborative learning in terms of individual-participant-single-modality, individual-participant-multiple-modality, and multiple-participant-multiple-modality. Analysis of annotated emotions, body gestures, and their interactions indicates that our multimodal dataset with designed treatments could effectively examine moments of regulation in collaborative learning. In addition, preliminary experiments based on baseline models suggest that the dataset provides a challenging in-the-wild scenario, which could further contribute to the fields of education and affective computing.
 
\end{abstract}



\maketitle

\section{Introduction}
Collaborative learning is a social system in which groups of learners solve problems or construct knowledge by working together \cite{dindar2022detecting}. Recent findings demonstrate that collaborative learning can promote higher-level thinking, oral communication, leadership skills, student-faculty interaction, and student responsibility \cite{sumtsova2018collaborative}. Although many factors can affect collaborative learning, social interaction has been considered one of the most important \cite{soller2001supporting,qureshi2021factors}. To succeed in collaboration, learners should actively exchange their ideas, experience, resources, skills, and feelings within a team \cite{sims2003promises,rimor2010complexity}. According to the research on promises of interactivity \cite{sims2003promises}, interactions enable collaborators to learn and encourage them to be focused, participative and dedicated to interchange ideas with each other. To this end, studying and promoting the interactions in a collaborative setting will provide valuable insight into the quality of collaboration and be significant and helpful in various fields, especially education research \cite{herrera2021collaborative}. 

There is a growing interest in studying interactions in a collaborative learning context by utilizing emotional and physiological measures in recent years \cite{dindar2020leaders,dindar2022detecting}. Thanks to the development of the hardware and AI technologies \cite{li2021deep,liu2022uncertain, li2015more,liu2021sg,li2016fast}, it is convenient to unobtrusively and automatically capture the physiological and visual signals of team members, which makes it possible to study the correspondence between multimodal signals for observing interactive processes during collaborative learning.

\begin{figure}
     \centering
     \begin{subfigure}[b]{0.45\textwidth}
         \centering
         \includegraphics[width=\textwidth]{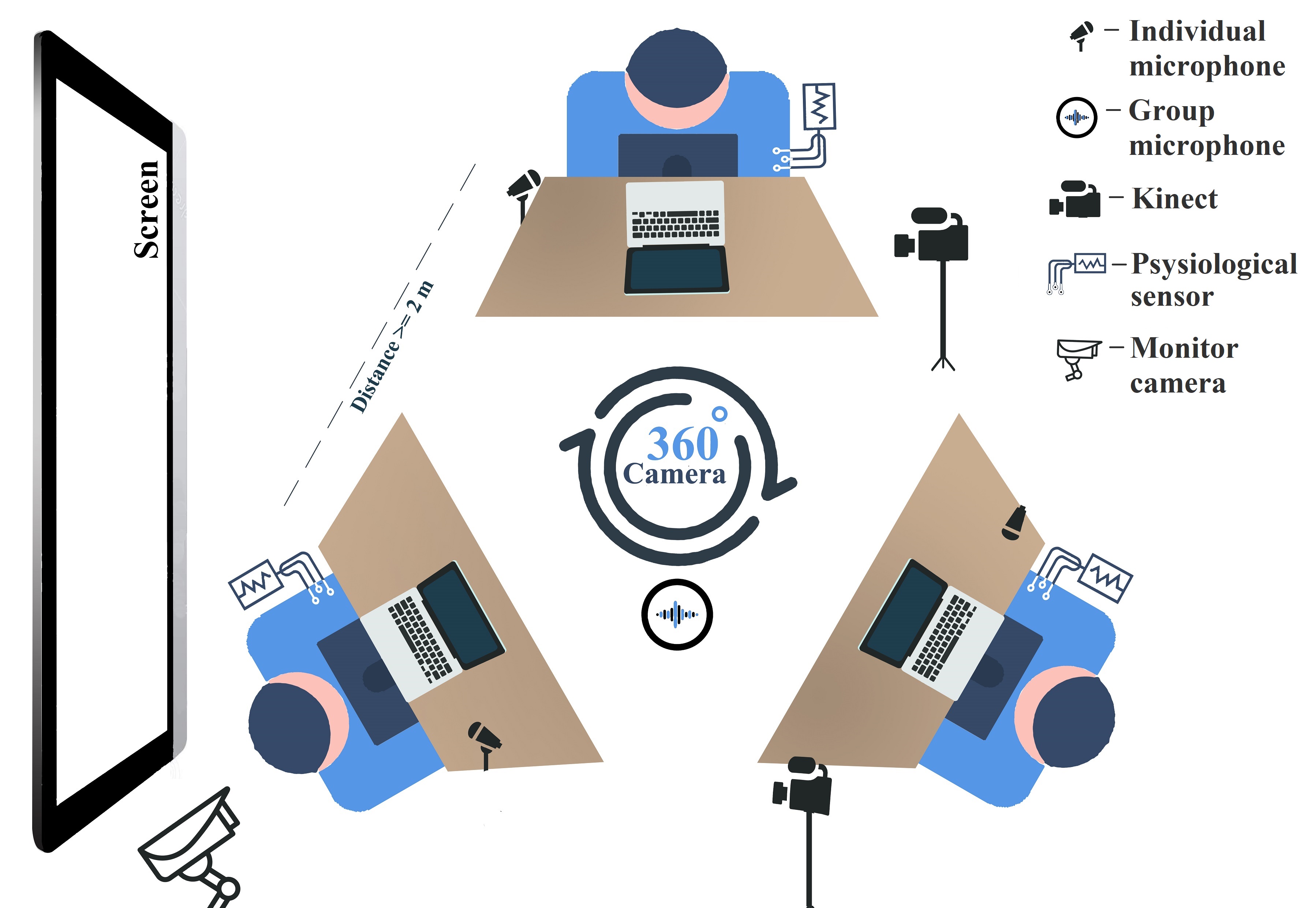}
         \caption{}
         \label{fig:demonstration}
     \end{subfigure}
     \hfill
     \begin{subfigure}[b]{0.47\textwidth}
         \centering
         \includegraphics[width=\textwidth]{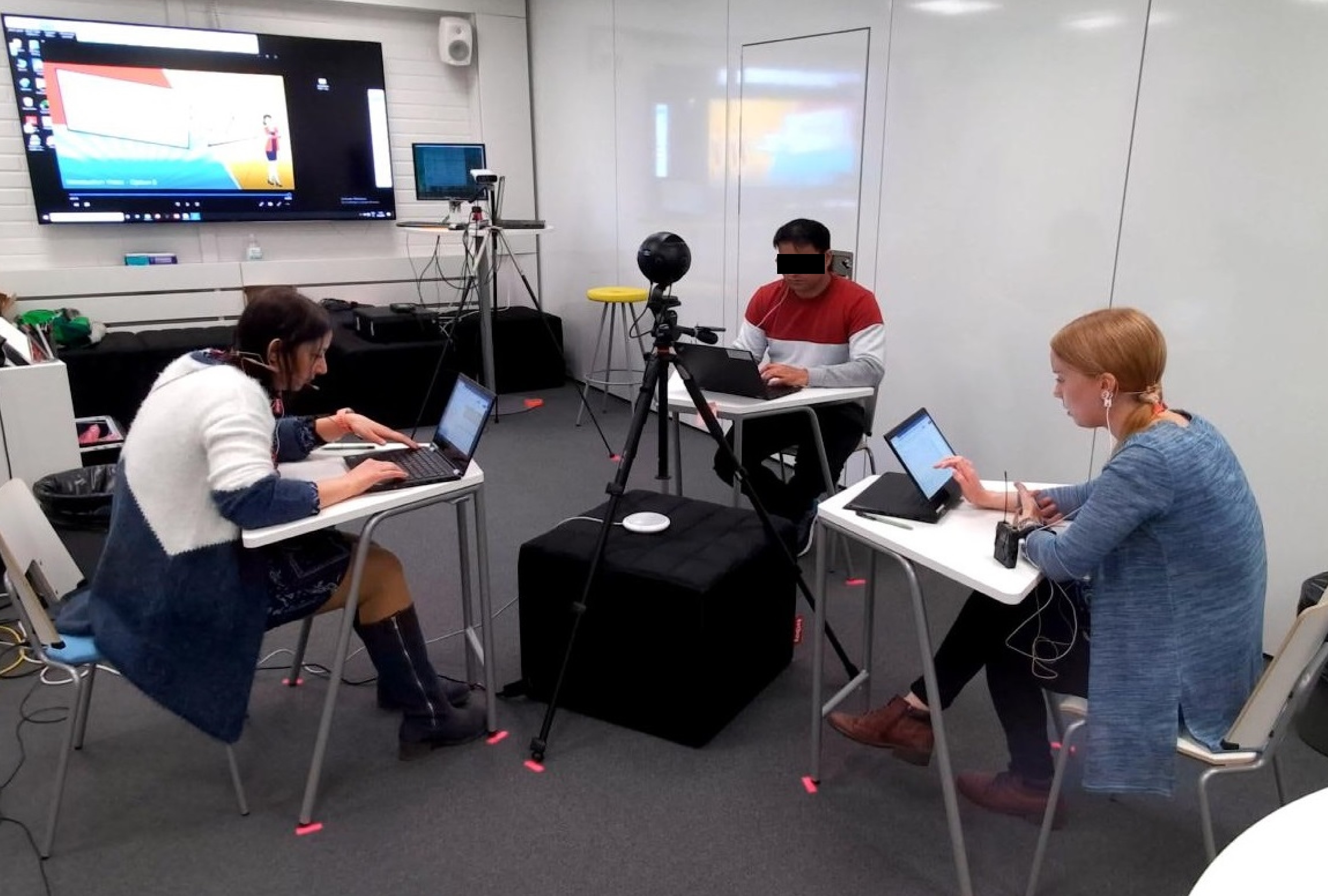}
         \caption{}
         \label{fig:real}
     \end{subfigure}
     \hfill
        \caption{The data collection setup. (a) An illustration of the seating plan and the location of the devices; (b) The real  environment of data collection.}
       \Description{The figures show the setting and environment of data collection.}
        \label{fig:settingposition}
\end{figure}

A variety of research has studied emotional interactions among learners \cite{dindar2020leaders,linnenbrink2011affect}. The results have revealed that positive emotional interactions are related to better collaboration. However, previous studies only focus on the facial expressions and the physiological signals in interactions independently. Many studies illustrate that the body gesture can also provide emotional clues \cite{liu2021imigue,liu2021graph}, and body gestures in interactions can contribute to solving cooperation and collaborative problem \cite{duarte2009gesture}. In this paper, we design a collaborative learning task and collect a multimodel dataset, including video, Kinect video, audio, and physiological data, to analyse collaborative learning interactions, as shown in Fig. \ref{fig:settingposition}. 

Another critical factor for successful learning is regulation \cite{jarvela2019capturing}. Socially shared regulation can promote productive collaborative learning. Thus, we introduce regulation processes in collaborative learning by designing interventions, i.e., the cognitive trigger and the emotional trigger in our task setting. Through analyzing different features such as emotional reflection during triggers, researchers can inspect whether and how those external events influence interactions of group members. 

In this paper, a new dataset is collected in terms of multiple modalities to comprehensively explore the regulation of learning and the aroused emotional interactions in collaborative learning. As far as we know, this is the first multimodal dataset for studying regulation in collaborative learning with regulatory triggers. The main contributions are as follows: 

\begin{itemize}
    \item We design a collaborative learning task with two kinds of triggers, i.e., cognitive and emotional triggers, to study the regulation inference in collaborative learning.
    \item We collect a multimodal dataset consisting of video, Kinect, audio, and physiological data to analyse  collaborative learning interctions.
    \item Statistical analysis and baseline experiments demonstrate that the regulation significantly impacts interactions during the collaborative process.
\end{itemize}


\section{Prior work}

\subsection{Collaborative learning}
Collaborative learning is an educational approach for enhanced learning \cite{herrera2021collaborative} where two or more learners work together to solve problems, complete tasks, or learn new concepts. Learners work as a group rather than individually to obtain a complete understanding by interchanging their ideas, processing and synthesizing information instead of using rote memorization of texts \cite{rimor2010complexity}. According to recent studies,  collaborative learning can boost higher-level thinking, oral communications, leadership skills, student-faculty interactions, self-esteem and responsibility of students \cite{laal2012benefits}.

Various interactions emerging in the collaborative learning process are essential features of effective learning. Many researchers have studied emotional interactions in collaborative learning.  Webb et al.  \cite{webb2013help} found that positive emotional interactions, like support and respect, are related to better collaboration. Dindar et al. \cite{dindar2020leaders} revealed that video-based facial emotion recognition helped explain social and affective dynamics in collaborative learning research. Besides emotions, physiological signals also serve crucial functions in studying collaborative processes \cite{dindar2022detecting}. Several studies have investigated the interaction in collaborative learning by identifying the synchronizing extent of physiological signals \cite{dindar2020leaders}. Additional work verified that  gestures served a variety of signalling functions in collaborative problem-solving communication and had a diagnostic role for team members \cite{reynolds2001gesture, duarte2009gesture}. 

Due to the contribution of different modalities, existing research has focused on studying the collaborative learning process by exploring multimodal signals \cite{nguyen2021multimodal,reilly2018exploring}. Nguyen et al. \cite{nguyen2021multimodal} developed a deep learning model to automatically detect interaction types for regulations in collaborative learning by analyzing electrodermal activities (EDA), video, and audio data. Reilly et al. \cite{reilly2018exploring} studied the Kinect and speech data and demonstrated how specific movements and gestures positively correlate with collaboration and learning gains. Unlike previous methods introducing only two or three modalities, in this paper, we collected a multimodal dataset including facial video, audio, gesture, EDA, heart rate (HR), and accelerometer to explore the collaborative process comprehensively. 


\subsection{Regulation in collaborative learning}

Recent research has highlighted the importance of co-regulation and socially shared regulation of learning to the group’s collaborative learning success \cite{malmberg2017capturing}. While self-regulated learning depicts the individual process of monitoring, reflecting, and correcting one’s emotion, motivation, and cognition towards attaining learning goals, co-regulation and socially shared regulation refer to this process in collaborative learning at the group level \cite{zimmerman2011handbook}. Co-regulation of learning relates to the co-operation of regulation in which self-regulated learning occurs with support from another learner. However, socially shared regulation of learning involves learners in a group interdependently regulating the group’s collaborative learning process and jointly regulating individual learning processes through social interactions \cite{hadwin2018self}. Therefore, besides multiple modalities, we also study regulation in collaborative learning tasks by introducing designed interventions.

\subsection{Relevant datasets}
There are currently various multimodal datasets used for studying emotion and gesture in collaborative learning. CMU Multimodal Opinion Sentiment and Emotion Intensity (CMU-MOSEI) dataset \cite{zadeh2018multi} contains more than 23,500 sentence utterance videos from online YouTube speakers. It has three modalities: language, visual, and acoustic, annotated for six basic emotions and five sentiments. The EmoReact Dataset \cite{nojavanasghari2016emoreact} is a multimodal emotion dataset of children which contains 1102 audio-visual clips of 17 different emotional states: six basic emotions, neutral, valence, and nine complex emotions, including uncertainty, curiosity, and frustration. Moreover, the Persuasive Opinion Multimedia (POM) corpus \cite{park2014computational} consists of 1,000 movie review videos obtained from a social multimedia website. This dataset includes three modalities, video, text, and acoustic, and annotates for multiple speaker traits. Although the above multimodal datasets are compatible for analyzing emotions in collaborative learning, they only consider individuals instead of interactions among multiple members in a group. Zhang et al. \cite{zhang2015learning} proposed a dataset studying the social relation between two or more people in one image or video. Alternatively, Kosti et al. \cite{kosti2019context} presented the EMOTIC dataset that involved scene context in addition to facial expression and body pose for extra information on emotion perception. However, these datasets only consider the visual modality and aim to study social or semantic relations. By contrast, our work establishes a multimodal dataset in collaborative learning scenario. It explores the interactions among group members and introduces regulations by designed interventions, which provides an interdisciplinary platform for studying emotion regulations and their impacts in collaborative learning.

\section{Dataset collection}
To systematically and comprehensively study the process of collaborative learning, we collect a multimodal dataset that contains facial videos, audio, physiological signals (including EDA, heart rate, and accelerometer), and Kinect data. This multimodal dataset is valuable for exploring:
\begin{itemize}
    \item Whether different modalities have underlying correlations in the collaborative process.
    \item Whether the fusion of various data sources could facilitate the task of collaborative learning.
    \item Whether the regulation has impacts on multiple modalities during the collaborative process.
\end{itemize}
Details of the data collection and annotation are explained in the following subsections.



\subsection{Equipment setup and data synchronization}
Our data recording was held in a laboratory studio, and the setup is shown in Fig. \ref{fig:settingposition} (a) and (b). Specifically, three participants sit in front of laptops. Two-meter COVID social distance was kept between participants during data collection. 

\begin{figure}
     \centering
     \begin{subfigure}[b]{\textwidth}
         \centering
         \includegraphics[width=\columnwidth]{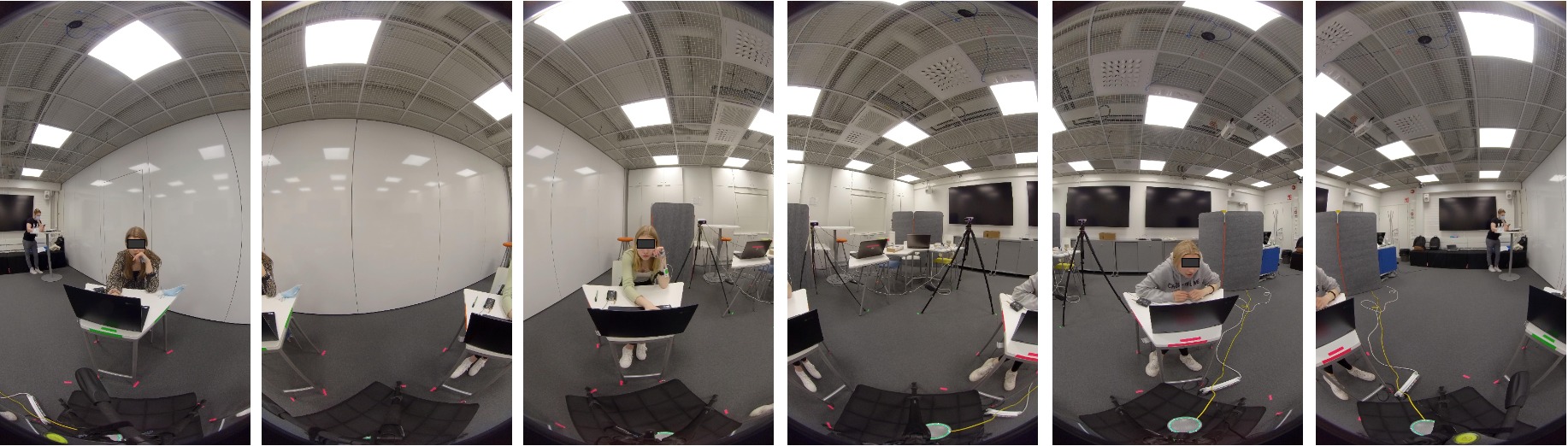}
         \caption{Example view of six cameras in  $360^{\circ}$ camera.}
         \label{fig:360_6}
     \end{subfigure}
     \hfill
     \begin{subfigure}[b]{\textwidth}
         \centering
          \includegraphics[width=\textwidth]{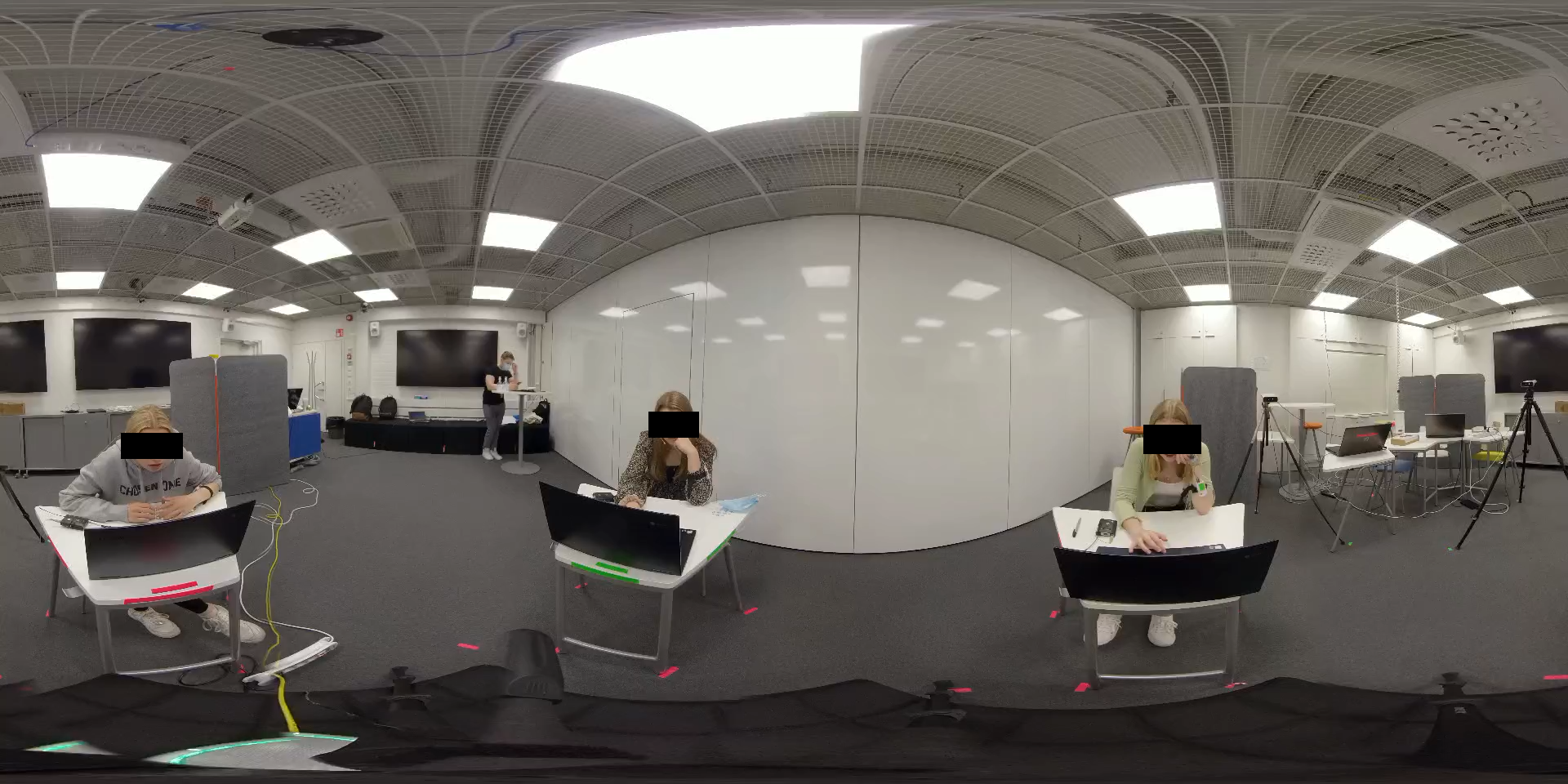}
         \caption{Example view of combined image in 360 degree.}
         \label{fig:360_1}
     \end{subfigure}
     \hfill
        \caption{Example of view of six cameras in  $360^{\circ}$ camera and the synthesized whole $360^{\circ}$ view. Zoom in for better view.}
        \label{fig:360}
\end{figure}

A $360^{\circ}$ camera (Insta360 Pro) that contains six camera spots and a microphone was placed in the center. The six cameras are hardware synchronized, and the grabbed frames from the six channels are used for building the whole environment in 360 degrees. During the collection, each participant was facing one camera directly. This way, we could have a compact frontal face for every participant, as shown in Fig. \ref{fig:360}. Resolutions of individual video and reconstructed video are 3840 x 2160 and 1920 x 960, respectively, with an average recording rate of 30 fps. Furthermore, a surveillance camera was applied to monitor and recall. Three individual microphones were employed to record the audio data.

\begin{figure}
     \centering
     \begin{subfigure}[b]{0.45\textwidth}
         \centering
         \includegraphics[width=\textwidth]{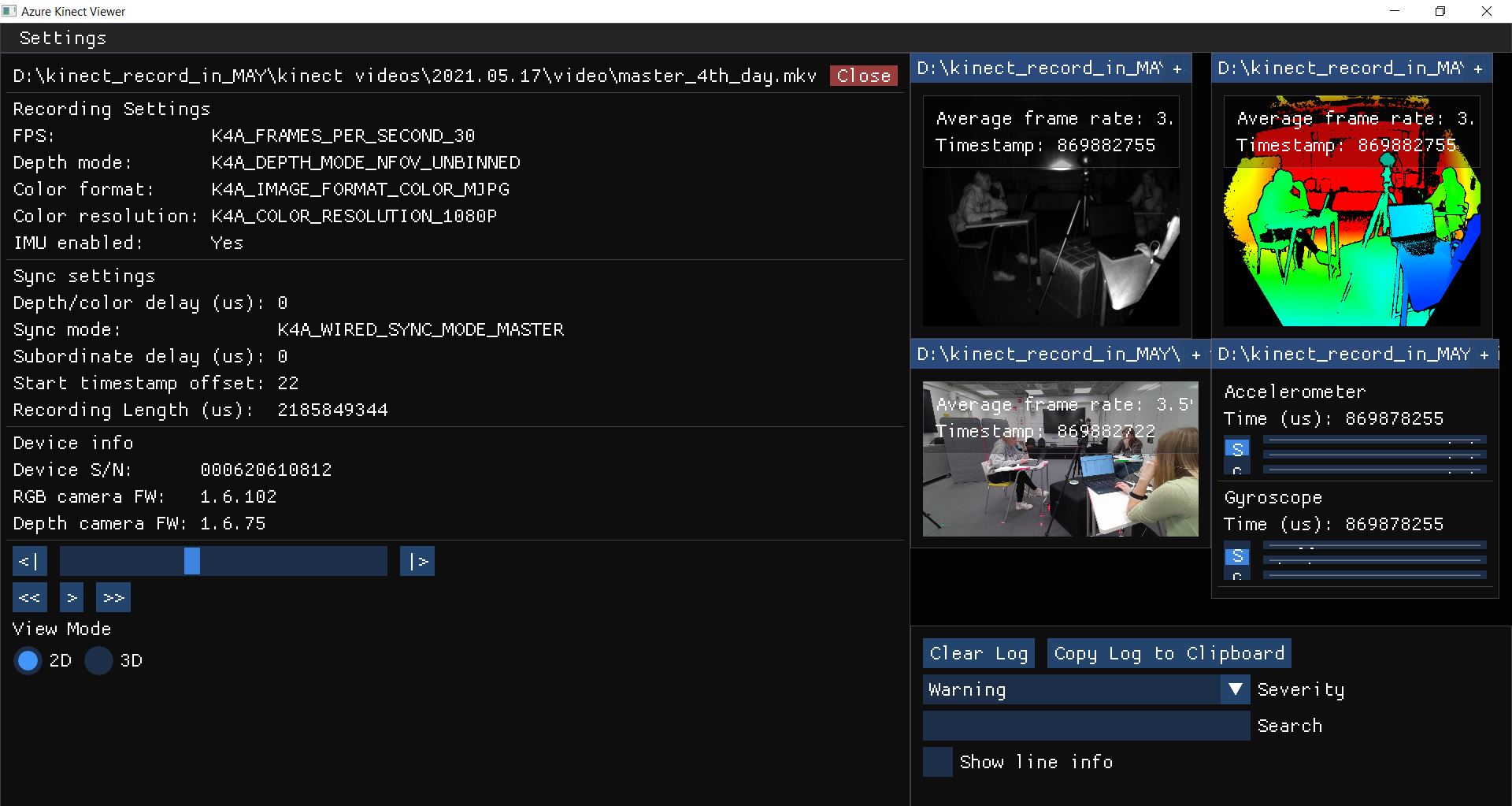}
         \caption{Kinect data recording and visualization}
         \label{fig:KD}
     \end{subfigure}
     \hfill
     \begin{subfigure}[b]{0.45\textwidth}
         \centering
         \includegraphics[width=\textwidth]{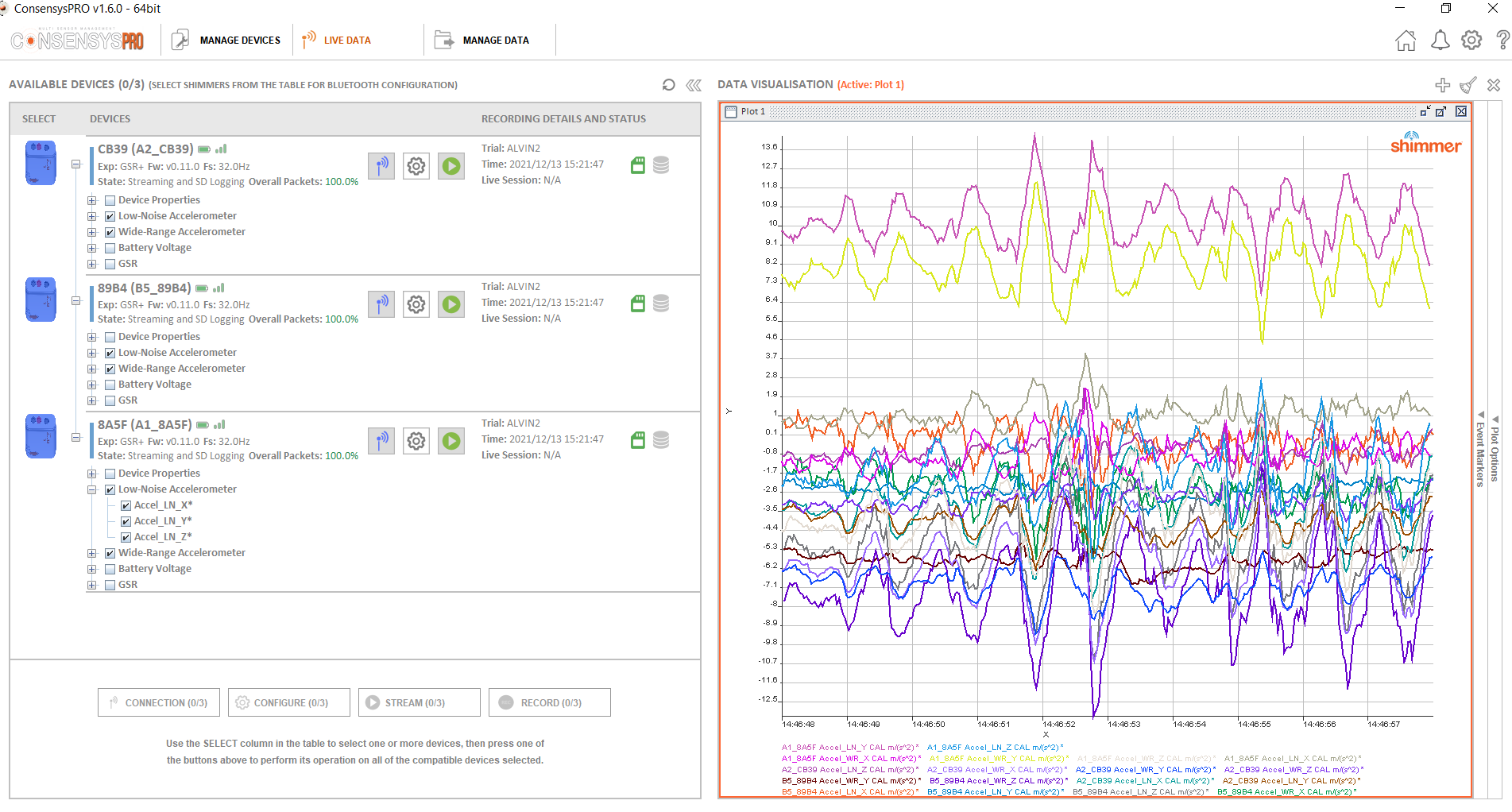}
         \caption{Physiological data recording and visualization}
         \label{fig:PD}
     \end{subfigure}
     \hfill
        \caption{Example of Kinect and physiological data with recording software. Zoom in for better view.}
        \label{fig:Sceen_Kinect_Pys}
        \Description{The figures (a) and (b) show the three modalities captured by Kinect camera and the accelerometer signal captured by Physiological sensors.}
\end{figure}

Two Kinect cameras (Azure Kinect DK) were utilized to capture the gesture of the three participants. The two devices were denoted as 'Master Kinect' and  'Salve Kinect' and synchronized by a cable automatically. Its average fps was around 30. Five sensor streams are aggregated in the Kinect camera, including a depth camera, a color camera, an infrared camera, IMU (Inertial Measurement Unit) and microphones. The Azure Kinect Viewer can visualize all the streams, as shown in Fig. \ref{fig:Sceen_Kinect_Pys} (a).

Physiological data, including EDA, HR, and accelerometer, were captured by Physiological sensors (Shimmer GSR3+) as shown in Fig. \ref{fig:Sceen_Kinect_Pys} (b). All the signals were collected at the sampling rate of 128 Hz, which could be used to reveal new insights into the emotional and cognitive processes in collaborative learning regulation \cite{dindar2022detecting}. 

Although the above multimodal data offers promising capabilities for analysis, the synchronization of multiple modalities collected from different channels is usually challenging in both methodological and theoretical aspects. In order to reach the finest granularity synchronization possible, the data synchronization was planned before the official collection in which each data collection device clock was synchronized to record Unix timestamp. The real-time timestamps were then used for data synchronization. The audio and video metadata were tracked with device-recorded Unix timestamps, while every record of physiological data was also associated with a specific Unix timestamp. Finally, the Kinect data was synchronized with physiological data by the frame change of the video played during the introduction of the collaborative tasks.

\subsection{Participants}
The study involved small groups of three high school students aged 15 years old in average (N=81, Male = 45, Female =36) who worked on a  collaborative task. The participants were recruited from high school classes through collaboration with the local teacher training school.  As a group, they worked on a shared Google document to design a healthy diet for a customer based on described nutritional needs. 


Specifically, the students were divided into 28 groups. Twenty-five groups were with three students. Three groups were only with two students. In order to study the external events' influence on collaborative learning, we designed the cognitive and emotional triggers in the middle of the learning procedure. During the entire collaborative process, the groups were divided into three types to evaluate the impacts of different triggers: Group A: without control (9 groups);  Group B: one cognitive trigger (9 groups); Group C: one cognitive trigger and three emotional triggers (10 groups).


The purpose and procedure of this research were explained to the students before the recording started. All students were aware that they could withdraw at any time of the collection. All students were asked to sign the consent form when he/she understood the contents and agreed to participate in the study.  Special questions were included in the consent form concerning data sharing related issues. Moreover, their guardians were informed about the study and also received GDPR document before the data collection.

\begin{figure}
    \centering
    \includegraphics[width=0.75\columnwidth]{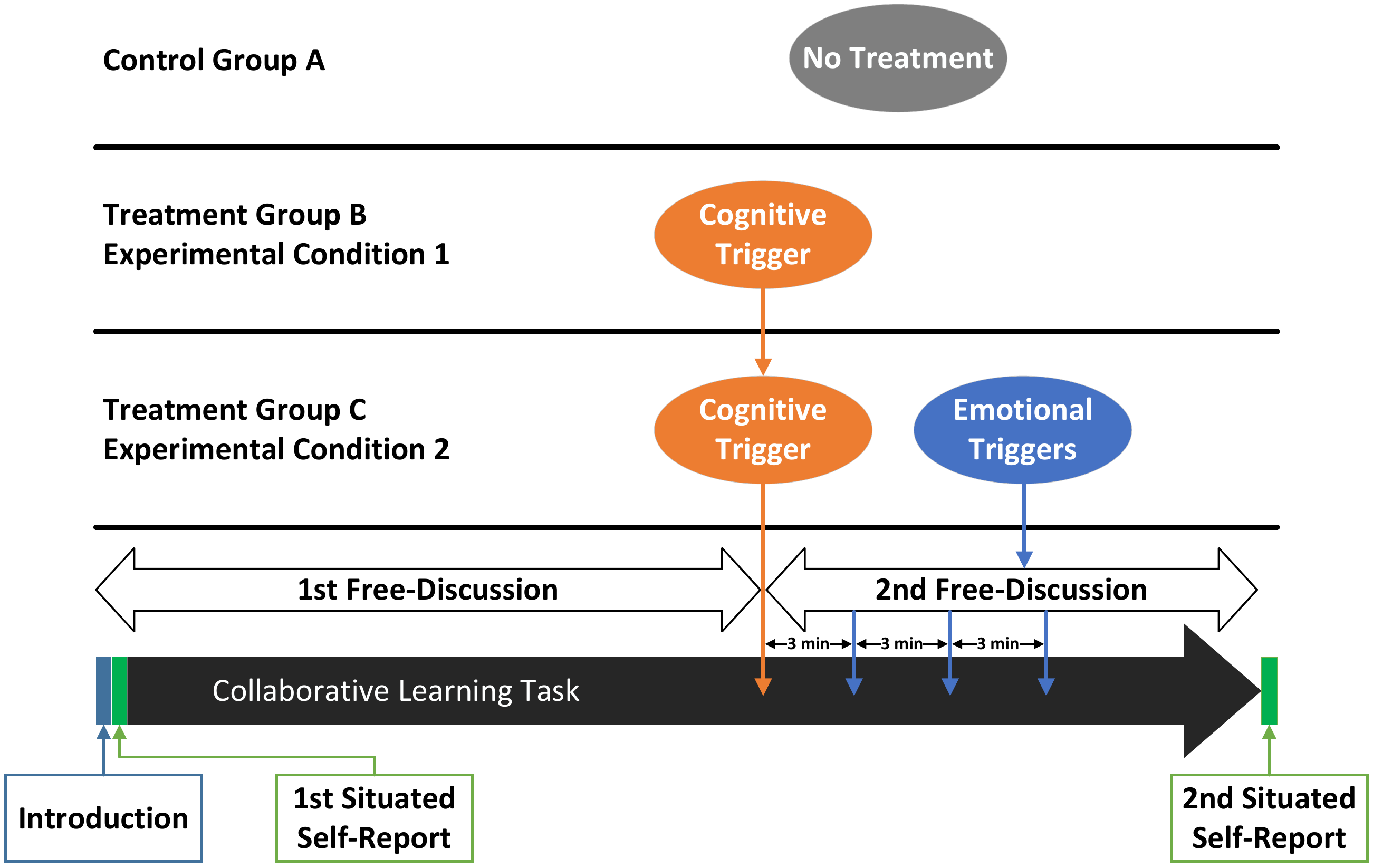}
    \caption{An illustration of the collaborative learning task. Group members work together to prepare a healthy smoothie for a customer. When the cognitive trigger applies, the customer says that she has an allergy to latex protein and dairy products. After that, the group will also be presented several emotional triggers at a specific time interval. Zoom in for a better view.}
     \Description{There are two phases in the collaborative learning task together with two kinds of triggers.}
    \label{fig:procedure}
\end{figure}

\subsection{Learning Procedure}
\begin{figure}
    \centering
    \includegraphics[width=0.85\columnwidth]{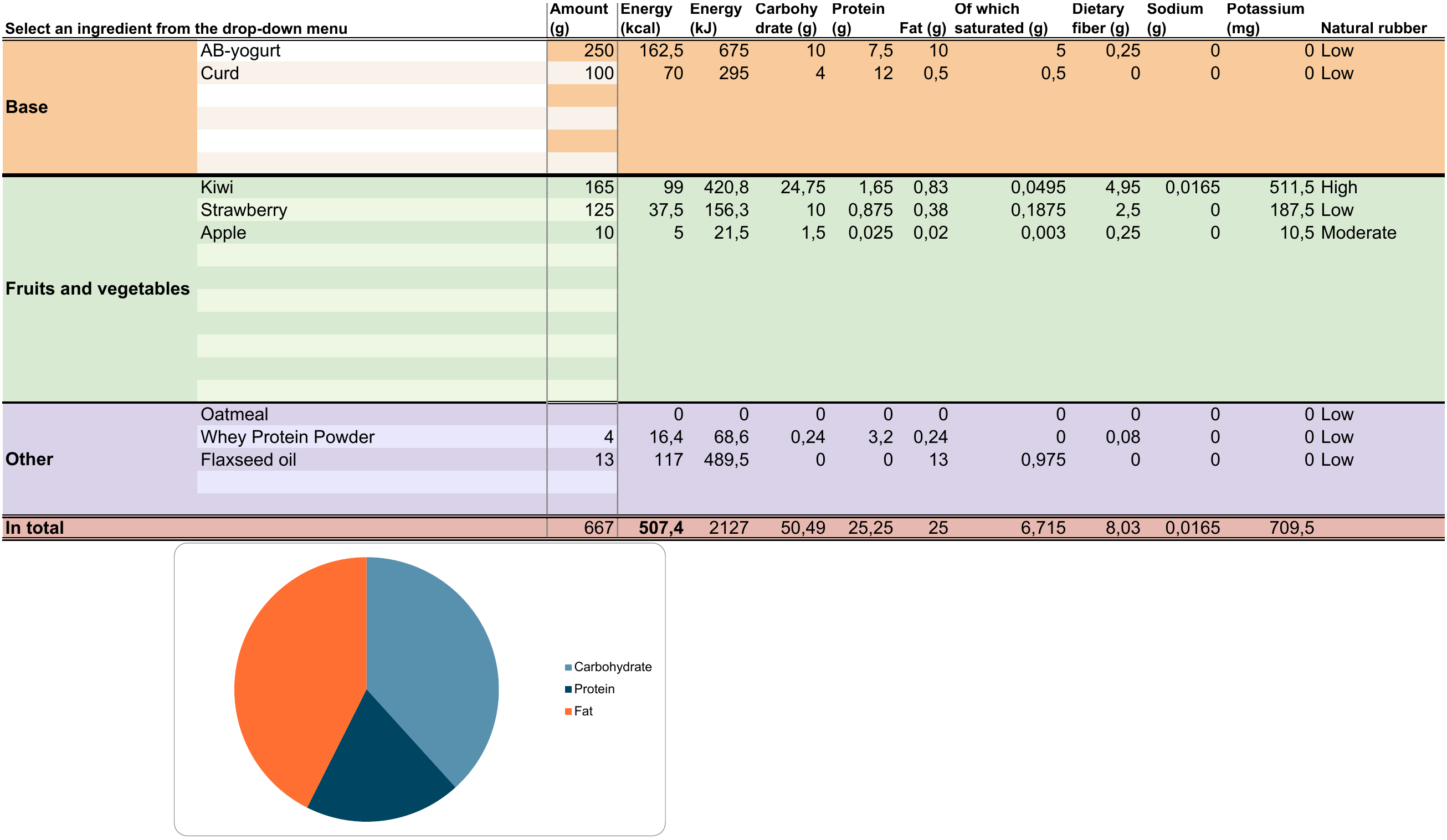}
    \caption{Example of participants' screen interfaces. The participants are required to make a healthy smoothie café.}
    \Description{The participants requires to make a healthy smoothie café.}
    \label{fig:screen}
\end{figure}

During the collection, participants would act as nutrition  specialists for a smoothie café. Their task was to plan a recipe for customers that supported the immune system during the pandemic. As shown in Fig. \ref{fig:procedure}, the learning procedure was divided into several phases in the following order: 
\begin{itemize}
    \item[i] INTRO (5 mins): An introduction video about general and practical issues was first played to help participants become familiar with the experimental setup and learning target.
    \item[ii] 1st SELF-REPORT (5 mins): Participants were asked to fill out pre-situated self-reports about their study habits, thoughts, and feelings.
    \item[iii] 1st FREE-DISCUSSION (15/25 mins): Participants worked on the meal plan via a shared Google Sheet document (see Fig. \ref{fig:screen}). Detailed task descriptions and all the needed materials could also be found in the document.
    \item[iv] 2nd FREE-DISCUSSION (15 mins): An cognitive trigger was imposed on Group B \& C, indicating that the coming customer had allergies or dietary restrictions. According to their working progress, one to three emotional triggers (every three minutes) indicating the customer's impatience were imposed on Group C three minutes after the cognitive trigger. 
    \item[v] 2nd SELF-REPORT (5-15 mins): All trigger moments were displayed to participants after they submitted the result. They were asked to recall their emotions during the triggers and fill out the post-situated self-reports.
\end{itemize}

One researcher was in the room with each group the whole time to ensure the smooth procedure of the experiment but would not be involved in the collaborative learning and answer any task-related questions. Smoothie vouchers were promised to motivate participants to engage in the learning task.

\subsection{Data annotation}
Based on the video, the data 30 seconds before and after every trigger were annotated with three emotion categories, including \textit{negative}, \textit{positive}, and \textit{neutral}. The annotation process was conducted in three steps. First, we extracted the frames around the cognitive and emotional triggers and roughly cropped the facial regions to make the facial expression easy to be followed. Second, ten annotators worked independently after a preparatory course. Each annotator was required to annotate three trigger clips. Labels were assigned to every trigger clip in seconds instead of frames because the emotion changes in an evolutionary manner. A tool was developed to play the frames in seconds for annotation continuously. Finally, we carried out fine-grained annotations by a 3-fold validation.

\subsection{Data statistics and samples }
Due to an unexpected hardware failure, the physiological data of nine participants and $360^{\circ}$ videos of three participants were lost. Eventually, the rest 78 participants’ data are complete and have been processed for analysis. Around 2730 minutes of frontal facial videos and audio data were recorded from 78 participants. Around 630 minutes of Kinect videos were collected from 30 participants of Group C with both cognitive and emotional triggers. Moreover, around 2040 minutes of physiological data were collected, including HR, EDA, and accelerometer. The data statics of multiple modalities are presented in Tab. \ref{tab:input}.

\begin{table}
    \centering
    \caption{Data statistics of multiple modalities}
    \label{tab:input}
    \setlength{\tabcolsep}{3mm}{
    \begin{tabular}{|l|l|l|l|l|}
        \bottomrule[1.5pt]
        \multicolumn{2}{|c|}{Modality} & \multicolumn{1}{c|}{Group} & \multicolumn{1}{c|}{Sample} & \multicolumn{1}{c|}{length/min} \\ \bottomrule[1pt]
        \multirow{2}{*}{Video} & Face & 23 & 78 & 2730 \\  \cline{2-5}
          & Kinect & 10 & 18 & 630\\ \bottomrule[1pt]
        \multicolumn{2}{|c|}{Audio} & 23 & 78 & 2730 \\ \bottomrule[1pt]
        \multirow{3}{*}{\makecell[c]{Physiological  signal}} & HR & \multirow{3}{*}{19} & \multirow{3}{*}{66} & \multirow{3}{*}{2310} \\  \cline{2-2}
          & EDA & & & \\ \cline{2-2}
          & Accelerometer & & & \\ 
        \bottomrule[1.5pt]
    \end{tabular}}
\end{table}

\section{Preprocessing}

\subsection{Emotion extraction from video data} \label{sec:baseline}
Two state-of-the-art emotion recognition methods proposed in 2021 are employed to fully use our 2D frontal video data, i.e., EmoNet \cite{toisoul2021estimation} and Emotion-GCN \cite{antoniadis2021exploiting}, which can extract discrete and continuous emotion simultaneously. Before feeding data into networks, face detection was firstly conducted with Dlib, a toolkit containing machine learning algorithms, to crop the facial region from collected videos. The emotion of each frame in all the videos were extracted by EmoNet and Emotion-GCN methods with cropped faces as inputs.

\subsubsection{Approach 1 - EmoNet} EmoNet \cite{toisoul2021estimation} was built on the top of the face-alignment network to predict fiducial landmarks, discrete emotional classes, and continuous affective dimensions on the face in a single pass. EmoNet provided two models pre-trained on the AffectNet dataset \cite{mollahosseini2017affectnet}. The two models are in five emotional classes (\textit{Neutral}, \textit{Happy}, \textit{Sad}, \textit{Surprise}, and \textit{Fear}) and eight emotional classes (\textit{Neutral}, \textit{Happy}, \textit{Sad}, \textit{Surprise}, \textit{Fear}, \textit{Anger}, and \textit{Contempt}), denoted as 'EmoNet5' and 'EmoNet8', respectively. The pre-trained models were evaluated on the cleaned AffectNet test set. The accuracy on five and eight emotions is $82\%$ and $75\%$, respectively.  The valence and arousal with models trained for five and eight emotions are 0.90 and 0.80, and 0.82 and 0.75 in terms of Concordance correlation coefficient (CCC), respectively \cite{crawford2007computer}. 

\subsubsection{Approach 2 - Emotion-GCN} Emotion-GCN \cite{antoniadis2021exploiting} is a multi-task learning framework that exploits dependencies between seven discrete emotional labels (\textit{Neutral}, \textit{Happy}, \textit{Sad}, \textit{Surprise}, \textit{Fear}, and \textit{Anger}) and two continuous affective dimensions (\textit{Valence} and \textit{Arousal}) using stacked Graph Convolutional Networks (GCNs) \cite{kipf2017semi}, which guide the representation learning of a backbone network for facial expression recognition in-the-wild. The Emotion-GCN was trained and evaluated on AffectNet and Aff-Wild2 \cite{kossaifi2017afew} datasets. The accuracy of the categorical model is $66.46\%$ and $48.92\%$, respectively. The CCC performance of valence and arousal on the dimensional model are 0.767 and 0.649, and 0.457 and 0.514, respectively.  

  
\subsection{Skeleton extraction from Kinect data}
\begin{figure}
     \centering
     \begin{subfigure}[b]{0.47\textwidth}
         \centering
         \includegraphics[width=\textwidth]{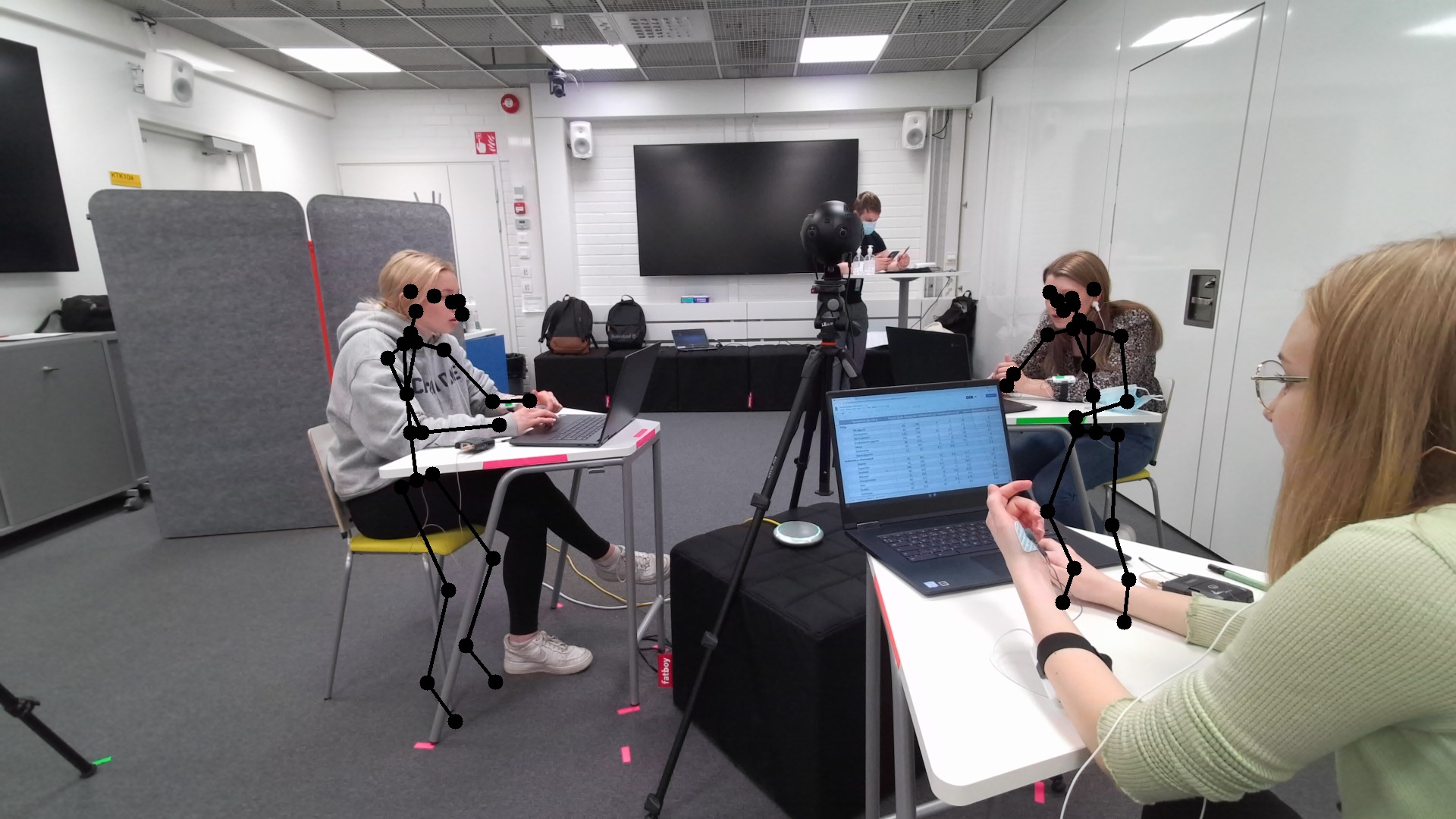}
         \caption{Master Kinect}
         \label{fig:master}
     \end{subfigure}
     \hfill
     \begin{subfigure}[b]{0.47\textwidth}
         \centering
         \includegraphics[width=\textwidth]{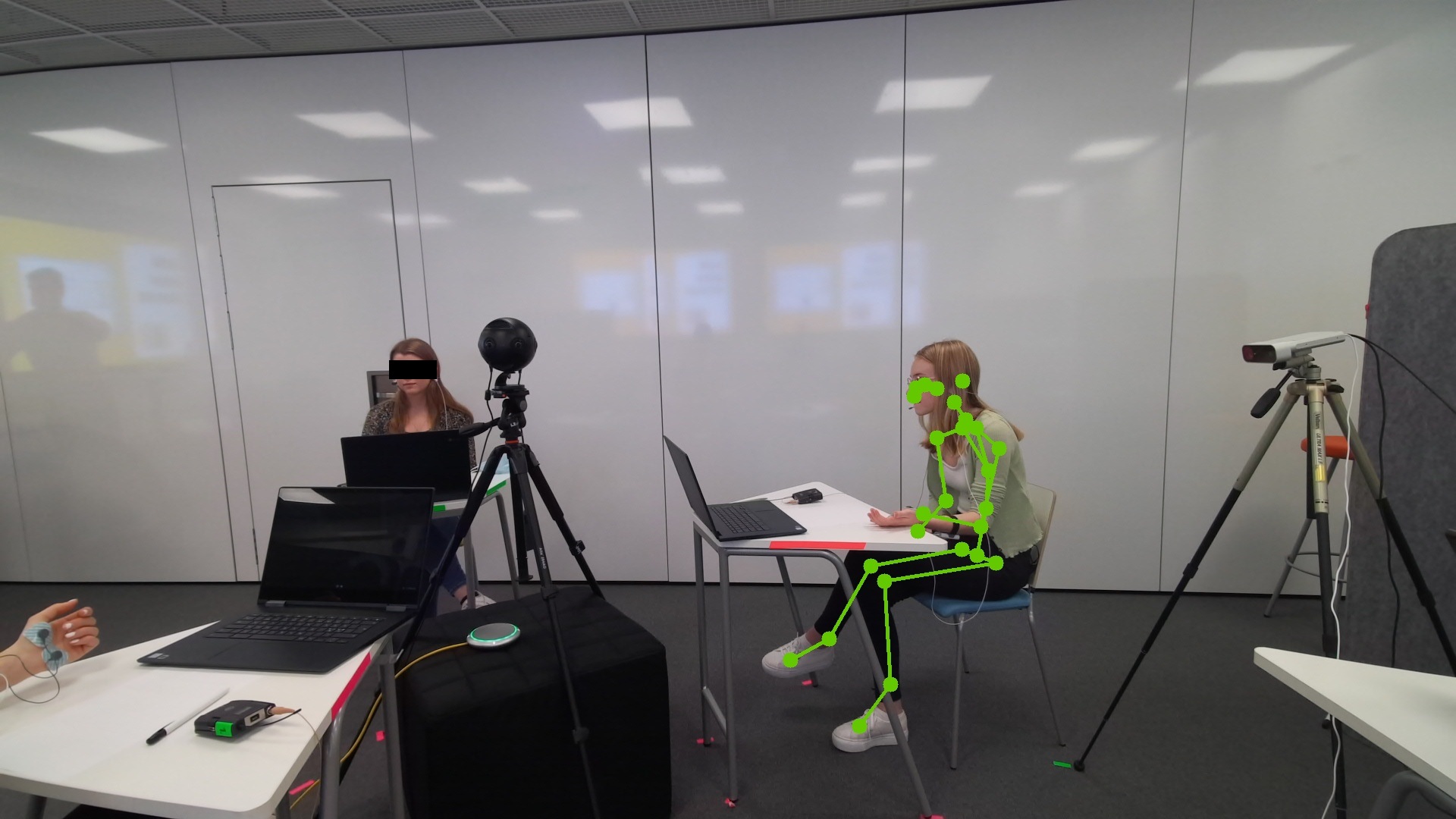}
         \caption{Slave Kinect}
         \label{fig:slave}
     \end{subfigure}
     \hfill
        \caption{Examples of the skeleton extracted from master and slave Kinect, respectively.}
            \Description{The skeleton can be extracted from videos.}
        \label{fig:skeleton_master_salve}
\end{figure}

The skeleton was extracted using the Kinect Azure devices, as shown in Fig. \ref{fig:skeleton_master_salve}. The engagement level of each subject is measured using the Euclidean distance of all joints moved within every second, which can be referred to as the summed movement speed of all joints. We calculated the joint movement speed during the extraction using three sets of joints, i.e., full-body joints with the head, upper body joints including the head, and upper body joints without the head. Besides, we also computed the leaning angle of the spine (to the forward direction) per second, which indicates how the subject is focusing.

\subsection{Physiological data}

The collected physiological data from Shimmer devices were extracted and then matched with associated learners in every group. Different data types were separated from the raw physiological data, including HR, EDA, and accelerometer. We conducted the decomposition to extract phasic and tonic features for EDA data. Furthermore, the trough-to-peak analysis using the Neuro2kit package was applied to detect distinct skin conductance responses that reflected emotional arousal moments.

\section{Data Analysis and results}
This section evaluates the effectiveness of our designed triggers and their impacts on interactions in terms of individual-participant-single-modality, individual-participant-multiple-modality, and multiple-participant-multiple-modality. Note that changes of physiological signals referring to triggers are easily overwhelmed by motion noises \cite{zhang2014troika, zhang2015photoplethysmography}, so we exclude them in this work. 


\subsection{\textbf{Emotion evaluation on the video data}}
\label{sec:Emotion_evaluation}

As introduced in Sec. \ref{sec:baseline}, we evaluate the three pre-trained models, EmoNet5, EmoNet8, and Emotion-GCN, using the annotated facial expressions from video data around triggers and providing baselines for the task of emotion recognition. 

\begin{figure}
     \centering
     \begin{subfigure}[b]{0.3\textwidth}
         \centering
         \includegraphics[width=\textwidth]{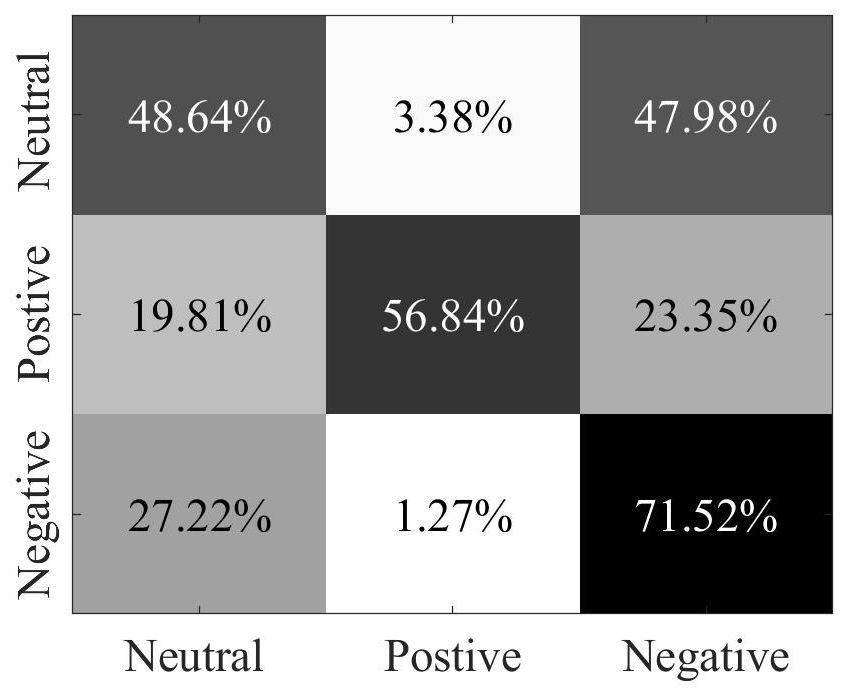}
         \caption{EmoNet5}
         \label{fig:ce5}
     \end{subfigure}
     \hfill
     \begin{subfigure}[b]{0.3\textwidth}
         \centering
         \includegraphics[width=\textwidth]{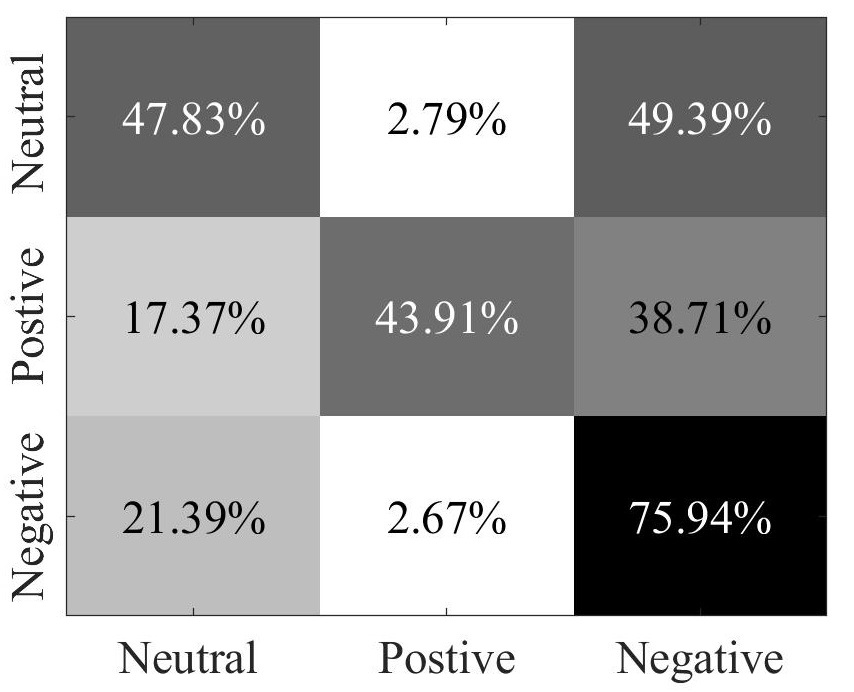}
         \caption{EmoNet8}
         \label{fig:ce8}
     \end{subfigure}
     \hfill
      \begin{subfigure}[b]{0.3\textwidth}
         \centering
         \includegraphics[width=\textwidth]{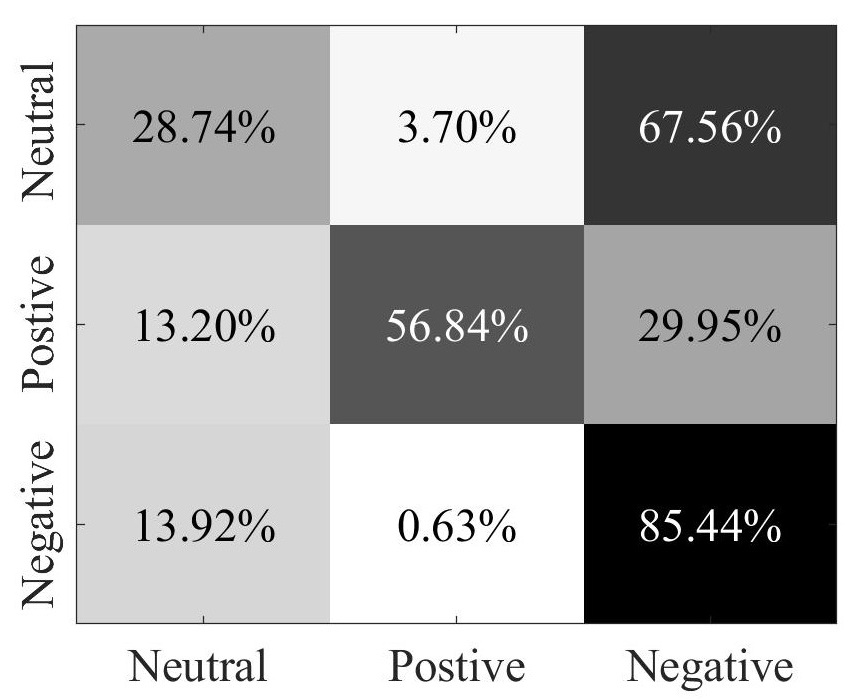}
         \caption{Emotion-GCN}
         \label{fig:ceGCN}
     \end{subfigure}
     \hfill
        \caption{The confusion matrices of the EmoNet5, EmoNet8, and Emotion-GCN on the annotated trigger clips.}
            \Description{The confusion matrices demonstrate the true labels and the predicted labels.}
        \label{fig:confusion_matrix}
\end{figure}

For discrete categories, since three emotion categories are considered during annotation, the emotion extracted by pre-trained models are mapped to \textit{Positive}, \textit{Negative}, and \textit{Neutral}. The \textit{Happy} is mapped to \textit{Positive}; the \textit{Sad}, \textit{Surprise}, \textit{Fear}, \textit{Anger}; and \textit{Contempt} are mapped to \textit{Negative}. In this case, the EmoNet5, EmoNet8, and Emotion-GCN achieve an accuracy of $51\%$, $48\%$, and $35\%$, respectively. The EmoNet outperforms Emotion-GCN, and the EmoNet5 achieves the best performance. One possible reason is that our dataset is in-the-wild and has various head poses and occlusions. The EmoNet aggregated face-alignment task can alleviate head poses and occlusions to some extent. Among the three emotions, we observe that \textit{Negative} gets the best performance, as shown in the confusion matrix in Fig.\ref{fig:confusion_matrix}, while \textit{Neutral} is easy to misclassify to \textit{Negative}. This phenomenon might be caused by significant face deformation because participants always look down at the laptop screen during the learning procedure. In general, the results suggest that the collected in-the-wild dataset of collaborative learning has head pose and face occlusion factors, which could be challenging in the facial expression recognition task.

In addition, We compare the valence with the emotion value reported in the self-reports to validate the continuous emotion. At the end of data collection, the participants were asked to watch the recorded video during the trigger clips and report their emotions when the trigger happened. Since they tend to report the emotions at the highest intensity after hearing the trigger, the predicted valence at the highest arousal after the trigger in 10s was compared with the self-reported valence. The average L1 norm distance between predicted and self-reported valence is 0.44, 0.44, and 0.46 on EmoNet5, EmoNet8, and Emotion-GCN. The results further indicate that the EmoNet based on face-alignment task have strength with the in-the-wild situation with various head poses and occlusions.

\subsection{ \textbf{Effect of triggers on emotion}}
\label{sec:effect_of_triggers_on_emotion}

\begin{figure*}
    \centering
    \includegraphics[width=0.9\columnwidth]{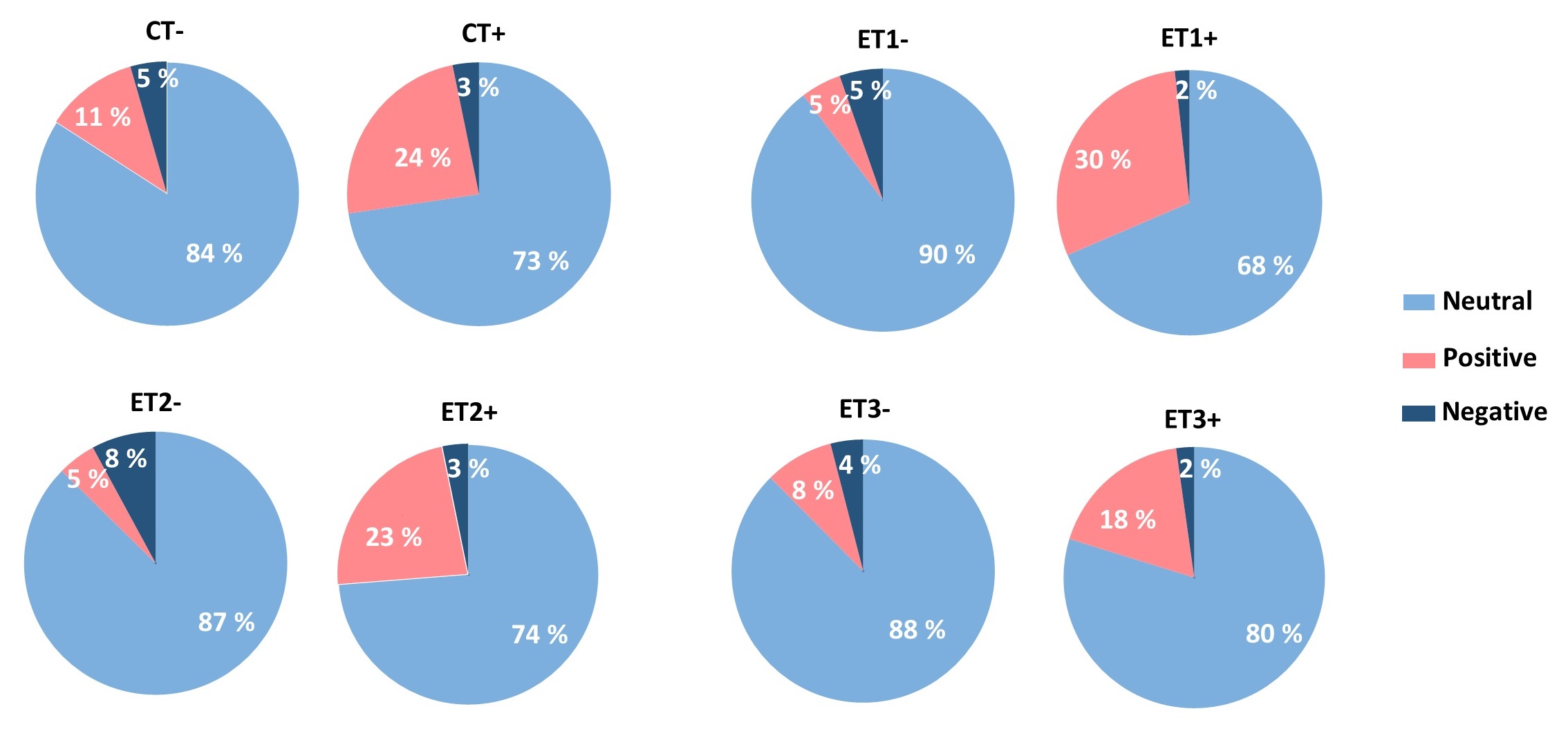}
    \caption{Emotion distribution before and after triggers. `-' and `+' represent before  and after trigger, respectively.}
        \Description{The triggers arouse positive emotion in this collaborative learning task.}
    \label{fig:EmotionDistribution}
\end{figure*}

The distribution of facial expressions before and after triggers in the 30 seconds is calculated in order to study the effect of the cognitive trigger (denoted as `CT') and emotional triggers (denoted as `ET') on emotion changes. From Fig. \ref{fig:EmotionDistribution}, it can be seen that the percentage of positive emotion increases when triggers activate. In particular, for the first emotional trigger 'ET1', the percentage of positive emotion is raised by $25\%$ compared with the emotion before the trigger, which is consistent with what was observed during the experiment. It means that the special request of the customer in CT and 'hurry up' in ETs did not arouse the participants' pressure or nervousness. One possible reason is that the participants understood it was just an experiment instead of a real scene of a smoothie store. In addition, they are teenagers in high school who are lively and laughing \cite{steinberg2014age}. Nevertheless, the results still illustrate that triggers impact emotions during collaborative learning. Chi-square results also confirmed the significant difference in the emotion distribution around different types of triggers, ${\chi}^2(6, N = 5940) = 27.52, \rho < .001.$

\subsection{\textbf{Emotion correlation between the participants}} 
\label{sec:Emotion_correlation}

This subsection investigates whether there is an association between the emotions of two participants during the trigger moments. Considering that the correlation between the paired emotions is not a linear relationship, we exploit Spearman’s correlation to assess the emotional relationship between the participants \cite{puth2015effective}. Spearman's rank-order correlation measures the strength and direction of the association/relationship between two continuous or ordinal variables \cite{ramsey1989critical}. The correlation coefficient ($r_s$) can take values from $+1$ to $-1$, which indicates a perfect positive ($+1$) or negative ($-1$) association of ranks. 

Tab. \ref{tab:Emotion_correlation} shows the correlation coefficient ($r_s$) and two-tailed significance value ($\rho$-value) of Spearman's rank-order correlation between paired emotions during different triggers. The duration of each trigger is 60 seconds. As there are three participants in every group, the emotions of three participants are concatenated to [$E_{P1_1}$,...$E_{P1_{60}}$,$E_{P2_1}$,...$E_{P2_{60}}$,$E_{P3_1}$,...$E_{P3_{60}}$] and [$E_{P2_1}$,...$E_{P2_{60}}$,$E_{P3_1}$,...$E_{P3_{60}}$,$E_{P1_1}$,...$E_{P1_{60}}$] to test the paired emotions. The total length for each trigger is 180. 
As presented in Tab. \ref{tab:Emotion_correlation}, in general, there is no negative emotion correlation between participants in one group. For specific triggers of some groups, such as CT and ET1 of G2, ET1 and ET3 of G7, and ET1 of G10, the $r_{s(178)}$ is more than 0.4 and $\rho<0.0005$. There was a statistically median positive correlation between the emotions in these groups. Especially for the ET2 of G7, there was a statistically significant, strong positive correlation between emotions during ET2 of G7, $r_{s(178)}= .728$, $\rho< .0005$. When we investigate the original emotions in these groups, it can be observed that the positive emotion occupied a more significant proportion compared with the other groups without statistically significant emotion correlation. The results suggested that positive emotion has a relationship with positive empathy, as described in the study of positive socioemotional interactions in collaborative learning \cite{linnenbrink2011affect}. In other words, positive socioemotional interactions lead to higher social regulation and better collaboration.

\begin{table}[t]
  \caption{Spearman's correlation between emotions by groups. The bold text represents median correlation, and the underlined text represents weak correlation. The bold and underlined text represents a strong correlation. `-' represents non-activated triggers, meaning the collaborative learning has already finished before the trigger. $r_s$ and $\rho$ represent the correlation coefficient and two-tailed significance value, respectively, which are also applicable in Table \ref{tab:skeleton_correlation}. The .000 represents the $\rho< 0.001$. }
  \centering
  \label{tab:Emotion_correlation}
  \begin{tabular}{lcccccccc}
      \bottomrule[1.5pt]
      Group & \multicolumn{2}{c}{CT} & \multicolumn{2}{c}{ET1} & \multicolumn{2}{c}{ET2} & \multicolumn{2}{c}{ET3} \\   \cline{2-9}
       & $r_s$ & $\rho$ & $r_s$ & $\rho$ & $r_s$ & $\rho$ & $r_s$ & $\rho$\\ \hline
      G1 & -.129 &.085&.116&.123&-&-&-&-\\
      G2&\textbf{.541}&\textbf{.000}&\textbf{.430}&\textbf{.000}&\underline{.289}&\underline{.000}&\underline{.243}&\underline{.000}\\
      G3&\underline{.298}&\underline{.000}&\underline{.345}&\underline{.000}&.051&.050&-.042&.577\\
      G4&.199&.007&\underline{.358}&\underline{.000}&-&-&-&-\\
      G5&.027&.716&\textbf{.463}&\textbf{.000}&-.039&.602&.027&.721\\
      G6&-.063&.405&\underline{.222}&\underline{.003}&\underline{.314}&\underline{.000}&-&-\\
      G7&.047&.528&\textbf{.420}&\textbf{.000}&\underline{\textbf{.728}}&\underline{\textbf{.000}}&\textbf{.461}&\textbf{.000}\\
      G8&-.029&.703&.042&.579&-.041&.590&--&--\\
      G9&\underline{.369}&\underline{.000}&\textbf{.458}&\textbf{.000}&\underline{.353}&\underline{.000}&-&-\\
      G10&\underline{.354} &\underline{.000}&\textbf{.435}&\textbf{.000}&\underline{.371}&\underline{.000}&-&-\\
      \bottomrule[1.5pt]
  \end{tabular}
\end{table}


\subsection{Skeleton analysis}
\label{sec:Skeleton_analysis}

To study the association of gestures between participants, we exploit Spearman’s correlation to assess the emotional relationship between skeleton moving speed extracted from the videos of master and slave Kinects. As shown in Tab. \ref{tab:skeleton_correlation}, there are more groups with a statistically significant, median positive correlation between the gesture speeds during `CT' compared with other triggers. One possible reason is that `CT' is the first trigger proposing the diet request leading to a stronger impression on the participants. There is no statistically significant, strong positive correlation in gesture speed between the skeleton in master and slave Kinects. A possible explanation for this might be the dynamic nature of regulation and interactions in collaborative learning \cite{jarvela2019capturing}. 

\begin{table}[h]
  \caption{Skeleton correlation between skeletons in master and slave Kinect. The bold text represents median correlation, and the underlined text represents weak correlation. * denotes the skeletons can not be extracted from the videos in G3 and G8 due to some hardware problem. `-' represents non-activated triggers, meaning the collaborative learning has already finished before the trigger. The .000 represents the $\rho< 0.001$.}
  \centering
  \label{tab:skeleton_correlation}
  \begin{tabular}{ccccccccc}
    \bottomrule[1.5pt]
    Group  &\multicolumn{2}{c}{CT}&\multicolumn{2}{c}{ET1}&\multicolumn{2}{c}{ET2}&\multicolumn{2}{c}{ET3}\\   \cline{2-9}
     &$r_s$ &$\rho$&$r_s$&$\rho$&$r_s$&$\rho$&$r_s$&$\rho$\\ \hline
    G1&.055&.675&.131&.317&-&-&-&-\\
    G2&\textbf{.428}&\textbf{.001}&.113&.389&-.008&.953&.053&.689\\
    G4&.090&.496&.080&.541&-&-&-&-\\
    G5&\textbf{.498}&\textbf{.000}&.065&.621&.181&.166&-.162&.216\\
    G6&\textbf{.455}&\textbf{.000}&.051&.699&\underline{.324}&\underline{.012}&-&-\\
    G7&\underline{.316}&\underline{.014}&.019&.886&-.045&.730&-.065&.620\\
    G9&-.036&.786&-.058&.659&.063&.634&-&-\\
    G10&.056 &.669&\underline{.232}&\underline{.075}&\underline{.261}&\underline{.044}&-&-\\
    \bottomrule[1.5pt]
  \end{tabular}
\end{table}

\subsection{Multimodal analysis}
\label{sec:multimodal_analysis}
\begin{figure}[h]
     \centering
     \begin{subfigure}[b]{\textwidth}
         \centering
         \includegraphics[width=\textwidth]{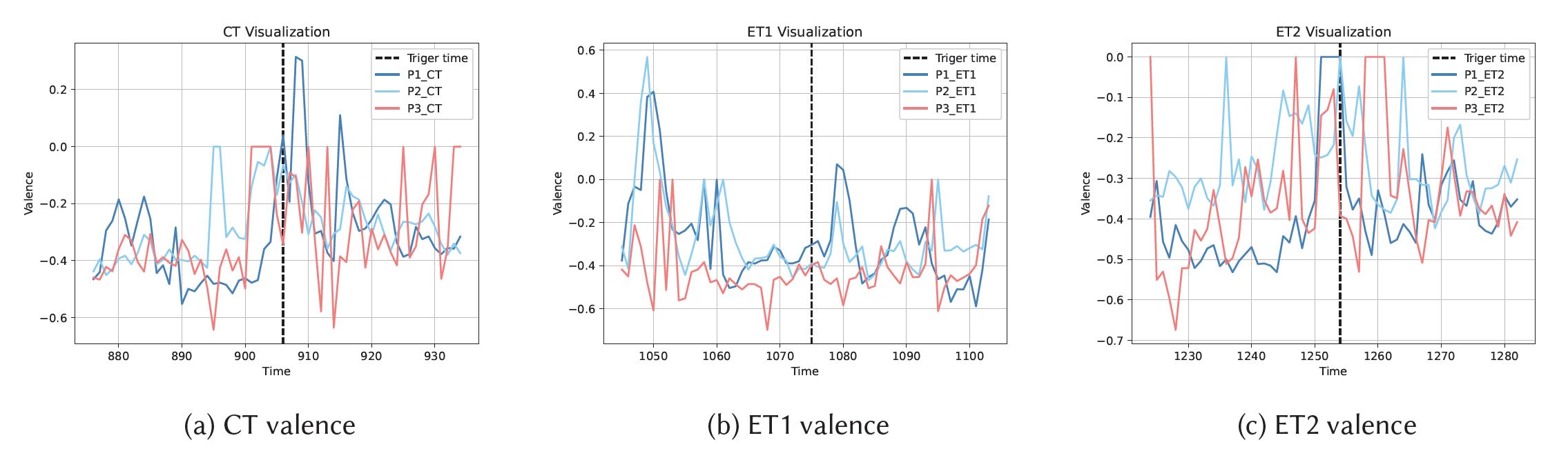}
         \label{fig:CT_valence}
     \end{subfigure}
     \hfill

     \hfill
     \begin{subfigure}[b]{\textwidth}
         \centering
 \includegraphics[width=\textwidth]{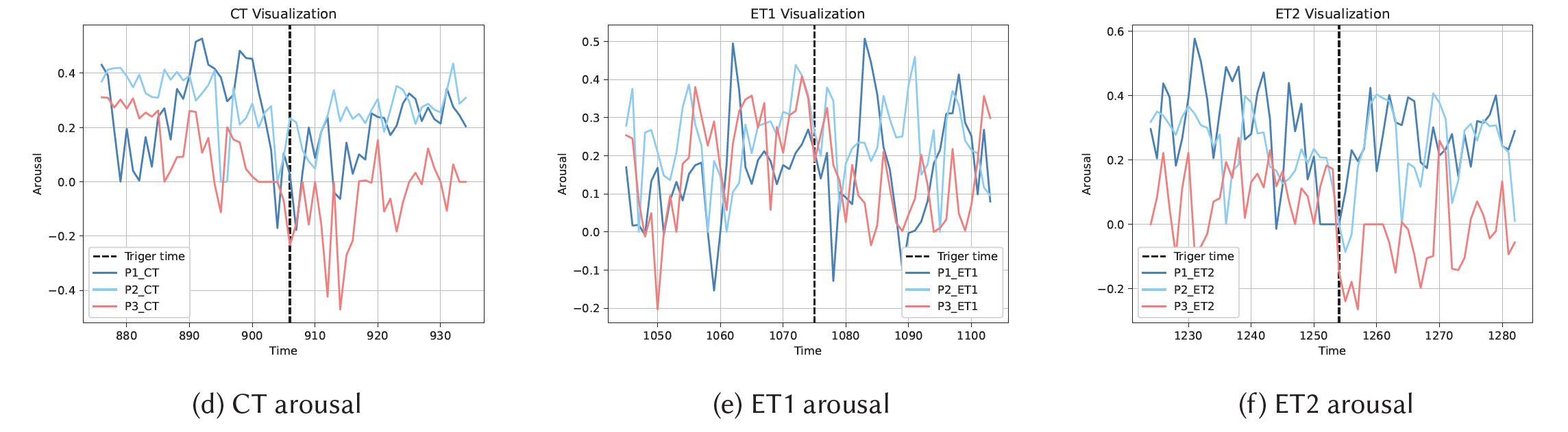}
         \label{fig:ET2_skeleton}
     \end{subfigure}

     \hfill
     \begin{subfigure}[b]{\textwidth}
         \centering
 \includegraphics[width=\textwidth]{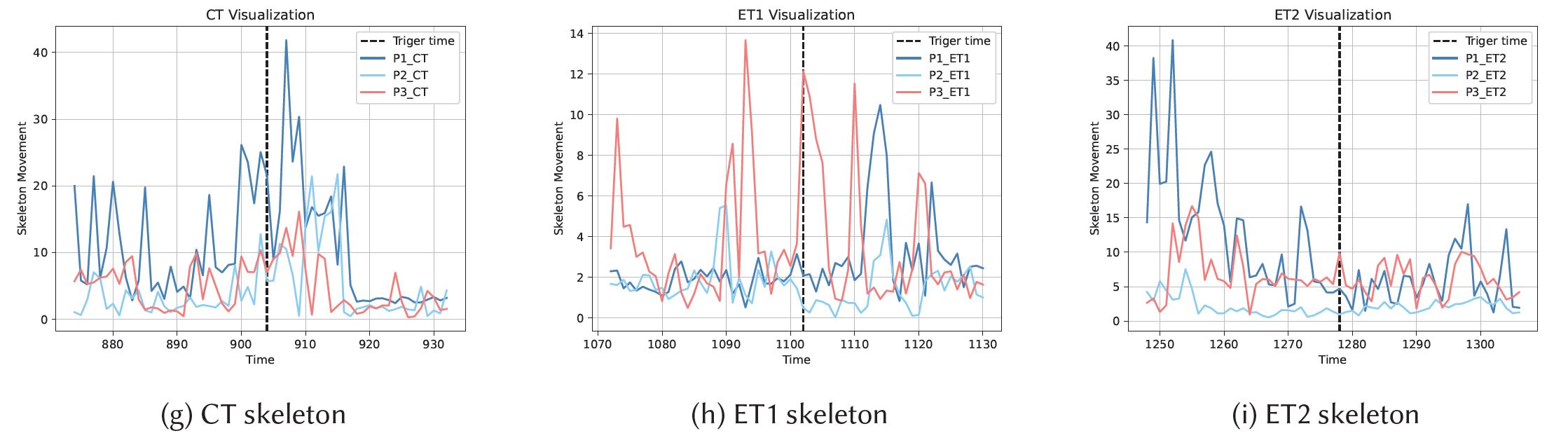}
         \label{fig:ET2_skeleton}
     \end{subfigure}

        \caption{Trend visualization of valence, arousal, and skeleton movement around cognitive and emotional triggers. Zoom in for a better view.}
        \Description{The visualization of multimodal data shows  there is an association between different modalities.}
        \label{fig:visualization}
\end{figure}

Apart from the single-modal analysis above, exploring whether there is an association between different modalities is exciting and necessary. To obtain a holistic view of all the trigger moments, we exploit continuous emotional states and body moving speeds around each trigger time-point and use data of G6 as an example for visualization. Fig. \ref{fig:visualization} shows valence, arousal, and skeleton movement trends of G6. For the valence, significant changes appear after applying every trigger, which confirms the result of Sec. 5.1. Situations of arousal are more complicated in this case. Inconsistent variations of each participant in different triggers further demonstrate the dynamic and the diversity of collaborative learning. The skeleton movement presents the characteristics of trigger-affected changes similar to the valence and shows that the effect of the same trigger decreases as the number of applications increases. In general, regardless of the modality or trigger, their trends correspond to previous correlations of emotion and skeleton. Similar consistency exists for other groups of this dataset, which demonstrates the potential value of our dataset in analyzing collaborative learning. 

\section{Discussion}
This paper introduces a novel multimodal dataset specifically designed to study regulation in collaborative learning. The dataset of collaborative learning groups contains 81 video clips from individual learners, which are annotated for three emotion labels around the intervention events, 28 360-stitched videos for learning groups, 18 Kinect depth sensor videos, and 66 128 Hz physiological signals. We respond to the recent calls to utilize multimodal data and advanced machine learning technologies to reveal the “unobservable” emotional and cognitive processes in collaborative learning and the induced regulation \cite{jarvela2020bridging,Nguyen2022exploring}. This paper also demonstrates a interdisciplinary approach with multimodal data to examine interactions for regulation in collaborative learning.   

Consistent with the literature \cite{nguyen2021multimodal,Nguyen2022exploring}, according to the data annotation and emotion evaluation on the video data, we found that facial expression recognition with in-the-wild interaction situations is challenging due to the head poses, occlusions, and non-emotional facial movements like talking \citep{liu2018visual}. As a result, novel methods such as more robust facial expression recognition algorithms are needed to be developed in the future. However, since learning regulation may rarely occur through interactions in collaborative learning \cite{noroozi2020multimodal}, existing datasets on collaborative learning would not be sufficient to develop such methods specifically for examining the regulation of learning. Accordingly, the contribution of this study has been to provide a multimodal dataset with designed interventions for regulation in collaborative learning. This dataset has significant implications for further methodological development and theoretical advancement in researching, understanding, and supporting regulation in collaborative learning. 
Another substantial contribution of this study relates to the interdisciplinary approach with preliminary results to examine the utility of the proposed dataset. Our results reported a significant difference in emotion distribution among different interventions to trigger regulation in collaborative learning. Our findings support previous learning sciences research of regulation in collaborative learning \cite{sobocinski2021exploring}. It demonstrates that the learning regulation is a temporal, cyclical, and dynamic process, and such external events would dynamically influence students' learning interactions. Our results also indicate that positive emotions are associated with positive empathy and better interaction. However, the gesture interaction has dynamic performance during different regulatory moments. In line with previous studies \cite{dindar2022detecting}, our results reveal that emotions aroused by the regulation of learning are displayed in multimodal emotional behaviors.

Our interdisciplinary approach in this presented study also responds to the recent calls for interdisciplinary effort bridging learning sciences, machine learning, and affective computing to maximize the impact of multimodal data and advanced techniques in examining and supporting regulation in collaborative learning (\cite{jarvela2020bridging, Nguyen2022exploring}. This paper contributes to both the field of computer sciences by offering a novel dataset for multimodal model development and to the field of learning sciences by providing new insights into the trigger moments for regulation in collaborative learning.

\section{Limitation and future work }
Since this is a preliminary study of multimodal analysis for collaborative learning, several limitations should be addressed in our future work. One thing is the insufficient annotated data. We only annotated the clips around triggers as a test set of the collected dataset for experiments in this paper. More labeled clips with discrete and dimensional emotions should be considered for a systematic analysis. Besides, instead of directly using pre-trained models, new methods need to be designed for robust emotion recognition in collaborative learning. 

On the other hand, the interaction analysis is mainly based on the correlation among individual facial expressions and gestures. High-level interaction annotations, such as eye contact, should be further considered. Another limitation is that this work only analyzes interactions influenced by triggers. The relationship between emotions and self-regulated learning should be further explored. Finally, multiple modalities, including facial behavior, physiological signal, and gestures, will be integrated and studied for a comprehensive understanding of collaborative learning. 


\newpage
\bibliographystyle{ACM-Reference-Format}
\bibliography{main}

\end{document}